\documentclass{article}

\PassOptionsToPackage{numbers,compress}{natbib}

\usepackage[dblblindworkshop, final]{neurips_2025}
\workshoptitle{The First Workshop on Generative and Protective AI for Content Creation}

\usepackage[utf8]{inputenc} 
\usepackage[T1]{fontenc}    
\usepackage{hyperref}       
\usepackage{url}            
\usepackage{booktabs}       
\usepackage{amsfonts}       
\usepackage{nicefrac}       
\usepackage{microtype}      
\usepackage{xcolor}         
\usepackage{graphicx}
\usepackage{colortbl}
\usepackage{multirow}
\usepackage{amsmath, amssymb}
\usepackage{subcaption}

\title{Hallucination as an Upper Bound: A New Perspective on Text-to-Image Evaluation}

\author{%
    Seyed Amir Kasaei \\
    Department of Computer Engineering,\\
    Sharif University of Technology \\
    \texttt{a.kasaei@me.com} \\
    \And
    Mohammad Hossein Rohban \\
    Department of Computer Engineering,\\
    Sharif University of Technology \\
    \texttt{rohban@sharif.edu} \\
}

\begin{document}
\maketitle

\begin{abstract}
In language and vision–language models, hallucination is broadly understood as content generated from a model’s prior knowledge or biases rather than from the given input. While this phenomenon has been studied in those domains, it has not been clearly framed for text-to-image (T2I) generative models. Existing evaluations mainly focus on alignment, checking whether prompt-specified elements appear, but overlook what the model generates beyond the prompt. We argue for defining hallucination in T2I as bias-driven deviations and propose a taxonomy with three categories: attribute, relation, and object hallucinations. This framing introduces an upper bound for evaluation and surfaces hidden biases, providing a foundation for richer assessment of T2I models.

\end{abstract}

\section{Introduction}

Hallucination, broadly defined as content generated from model priors rather than grounded in the input, is a critical challenge in contemporary AI. In large language models (LLMs) and vision--language models (VLMs), hallucination has been extensively studied because it undermines trust, factuality, and reliability, leading to a rich body of surveys and evaluation methods~\cite{huang2025survey,bai2024hallucination,chen2025survey,liu2024survey,manakul2023selfcheckgpt,guan2024hallusionbench,li2023halueval,chen2024diahalu,kaul2024throne}.  

In contrast, hallucination in text-to-image (T2I) generative models has not been clearly framed. Current evaluations overwhelmingly treat the problem as one of \emph{alignment}: verifying whether the objects, attributes, and relations explicitly mentioned in a prompt are faithfully represented in the generated image~\cite{qin2024evaluating,chen2025survey,hu2023tifa,ghosh2023geneval,huang2023t2icompbench,li2024vqascore,lim2025ihalla}. While these approaches have advanced prompt fidelity evaluation, they still capture only a \emph{lower bound} of performance, focusing on what is requested, but overlooking a complementary dimension: \emph{what does the model generates beyond the prompt?}  

In this position paper, we propose a clearer framing of hallucination in T2I generation: rather than equating it with prompt misalignment, we define it as bias-driven deviations that manifest through unintended attributes, relations, or objects. This distinction is essential because alignment metrics establish only a \emph{lower bound}, measuring whether requested elements are present, while hallucination evaluation sets an \emph{upper bound}, capturing what the model introduces beyond the prompt. By explicitly recognizing hallucination as this complementary dimension, we expose hidden biases that remain invisible to existing evaluations and provide a principled foundation for more comprehensive and reliable assessment of T2I models.  
\section{Hallucination Taxonomy and Evaluation Perspective}

We classify hallucinations in text-to-image generation into three types—object, attribute, and relation—based on how the model introduces unintended content. These hallucinations are distinct from alignment errors: instead of failing to follow instructions, the model extends the prompt beyond what the user explicitly requested. Each type raises unique challenges for evaluation and interpretability.

\subsection{Object hallucination}  
Text-to-image models frequently introduce entities not mentioned in the prompt. We refer to this behavior as \emph{object hallucination}. Unlike object neglect—where expected objects are omitted—hallucination involves generating new objects, driven by internal priors rather than user intent. Although these additions may be plausible, they often distort the focus or semantics of the prompt.

For example, a prompt like ``a bowl of apples'' might yield a bowl containing both apples and oranges; ``a horse'' might consistently appear with a rider; and ``a street with cars'' may include pedestrians or bicycles. These additions reflect model assumptions about scene completion, even when no such context was provided.

Formally, let a prompt $P$ specify a set of objects $O = \{o_1, o_2, \dots, o_n\}$. If the generated image contains a non-empty set $O'$ such that $O' \cap O = \emptyset$, then $O'$ constitutes object hallucination.

\subsection{Attribute hallucination}
Even when users omit attributes in their prompts, current text-to-image models often make specific visual assumptions. Rather than staying neutral, they tend to assign properties such as color, gender, style, or emotional tone by default. We define this behavior as \emph{attribute hallucination}, distinguishing it from attribute misalignment, where the model misrenders a requested detail.

For instance, the prompt ``a doctor'' may consistently produce a male figure in a white coat, despite no gender or clothing instructions. A request for ``a wedding cake'' may always yield a tall, tiered, white cake, implicitly enforcing one cultural template. Similarly, ``a child'' might be shown smiling outdoors in polished clothing, reflecting idealized emotional defaults. These decisions, while plausible, reflect unprompted biases that reduce diversity and interpretive openness.

Let $P$ refer to a prompt containing an object $o$ with no explicit attributes. If the image includes an attribute $a'$ not implied by $P$, we consider $a'$ to be an instance of attribute hallucination.

\subsection{Relation hallucination}
Models are also prone to inserting relationships between objects—even when no such connection is described in the prompt. This behavior, which we term \emph{relation hallucination}, introduces spatial or functional interactions that are not grounded in user input. It is distinct from relation misalignment, where a specified interaction is rendered incorrectly.

Such hallucinations may appear as default layouts or stereotyped activities. For example, ``a man and a dog'' may consistently depict the man walking the dog on a leash, implying control. A prompt like ``a woman and a laptop'' might always show her typing, suggesting a work scenario. Likewise, ``a child and a book'' often yields an image of the child reading, embedding an unintended learning narrative. These are not errors per se, but overlearned associations that compromise prompt neutrality.

Let $P$ contain a set of objects $O = \{o_1, o_2\}$ with no explicit relation. If the resulting image includes a relation $r$ not entailed by $P$, we categorize $r$ as relation hallucination.

\section{Conclusion}
Our taxonomy reframes hallucination in text-to-image generation as a distinct, complementary dimension to prompt alignment. By identifying how models inject unintended objects, attributes, or relations, we highlight the need to move beyond alignment-based lower bounds and evaluate upper-bound behavior. Hallucination evaluation reveals model biases that undermine controllability, neutrality, and trust—factors critical for real-world deployment. We hope this perspective motivates new benchmarks that explicitly measure what models generate \emph{beyond the prompt}.

\newpage
{\small
\bibliographystyle{abbrvnat} 
\bibliography{ref}

@article{huang2025survey,
  title={A survey on hallucination in large language models: Principles, taxonomy, challenges, and open questions},
  author={Huang, Lei and Yu, Weijiang and Ma, Weitao and Zhong, Weihong and Feng, Zhangyin and Wang, Haotian and Chen, Qianglong and Peng, Weihua and Feng, Xiaocheng and Qin, Bing and others},
  journal={ACM Transactions on Information Systems},
  volume={43},
  number={2},
  pages={1--55},
  year={2025},
  publisher={ACM New York, NY}
}

@article{bai2024hallucination,
  title={Hallucination of multimodal large language models: A survey},
  author={Bai, Zechen and Wang, Pichao and Xiao, Tianjun and He, Tong and Han, Zongbo and Zhang, Zheng and Shou, Mike Zheng},
  journal={arXiv preprint arXiv:2404.18930},
  year={2024}
}

@article{chen2025survey,
  title={A Survey of Multimodal Hallucination Evaluation and Detection},
  author={Chen, Zhiyuan and Min, Yuecong and Zhang, Jie and Yan, Bei and Wang, Jiahao and Wang, Xiaozhen and Shan, Shiguang},
  journal={arXiv preprint arXiv:2507.19024},
  year={2025}
}

@article{liu2024survey,
  title={A survey on hallucination in large vision-language models},
  author={Liu, Hanchao and Xue, Wenyuan and Chen, Yifei and Chen, Dapeng and Zhao, Xiutian and Wang, Ke and Hou, Liping and Li, Rongjun and Peng, Wei},
  journal={arXiv preprint arXiv:2402.00253},
  year={2024}
}

@article{manakul2023selfcheckgpt,
  title={Selfcheckgpt: Zero-resource black-box hallucination detection for generative large language models},
  author={Manakul, Potsawee and Liusie, Adian and Gales, Mark JF},
  journal={arXiv preprint arXiv:2303.08896},
  year={2023}
}

@inproceedings{guan2024hallusionbench,
  title={Hallusionbench: an advanced diagnostic suite for entangled language hallucination and visual illusion in large vision-language models},
  author={Guan, Tianrui and Liu, Fuxiao and Wu, Xiyang and Xian, Ruiqi and Li, Zongxia and Liu, Xiaoyu and Wang, Xijun and Chen, Lichang and Huang, Furong and Yacoob, Yaser and others},
  booktitle={Proceedings of the IEEE/CVF Conference on Computer Vision and Pattern Recognition},
  pages={14375--14385},
  year={2024}
}

@article{li2023halueval,
  title={Halueval: A large-scale hallucination evaluation benchmark for large language models},
  author={Li, Junyi and Cheng, Xiaoxue and Zhao, Wayne Xin and Nie, Jian-Yun and Wen, Ji-Rong},
  journal={arXiv preprint arXiv:2305.11747},
  year={2023}
}

@article{chen2024diahalu,
  title={Diahalu: A dialogue-level hallucination evaluation benchmark for large language models},
  author={Chen, Kedi and Chen, Qin and Zhou, Jie and He, Yishen and He, Liang},
  journal={arXiv preprint arXiv:2403.00896},
  year={2024}
}

@inproceedings{kaul2024throne,
  title={Throne: An object-based hallucination benchmark for the free-form generations of large vision-language models},
  author={Kaul, Prannay and Li, Zhizhong and Yang, Hao and Dukler, Yonatan and Swaminathan, Ashwin and Taylor, CJ and Soatto, Stefano},
  booktitle={Proceedings of the IEEE/CVF Conference on Computer Vision and Pattern Recognition},
  pages={27228--27238},
  year={2024}
}

@article{qin2024evaluating,
  title={Evaluating Hallucination in Text-to-Image Diffusion Models with Scene-Graph based Question-Answering Agent},
  author={Qin, Ziyuan and Cheng, Dongjie and Wang, Haoyu and Yi, Huahui and Shao, Yuting and Fan, Zhiyuan and Li, Kang and Lao, Qicheng},
  journal={arXiv preprint arXiv:2412.05722},
  year={2024}
}

@inproceedings{hu2023tifa,
  title={TIFA: Accurate and Interpretable Text-to-Image Faithfulness Evaluation with Question Answering},
  author={Hu, Yushi and Liu, Benlin and Kasai, Jungo and Wang, Yizhong and Ostendorf, Mari and Krishna, Ranjay and Smith, Noah A.},
  booktitle={Proceedings of the IEEE/CVF International Conference on Computer Vision (ICCV)},
  pages={20030--20041},
  year={2023}
}

@inproceedings{ghosh2023geneval,
  title={GenEval: An Object-Focused Framework for Evaluating Text-to-Image Alignment},
  author={Ghosh, Dhruba and Hajishirzi, Hannaneh and Schmidt, Ludwig},
  booktitle={Advances in Neural Information Processing Systems (NeurIPS)},
  year={2023}
}

@inproceedings{huang2023t2icompbench,
  title={T2I-CompBench: A Comprehensive Benchmark for Compositional Text-to-Image Generation},
  author={Huang, Kaiyi and Sun, Kaiyue and Xie, Enze and Li, Zhenguo and Liu, Xihui},
  booktitle={Advances in Neural Information Processing Systems (NeurIPS)},
  year={2023}
}

@inproceedings{li2024vqascore,
  title={VQAScore: A Vision-Language QA Approach for Automatic Text-to-Image Alignment Evaluation},
  author={Li, Baiqi and Lin, Zhiqiu and Pathak, Deepak and others},
  booktitle={Proceedings of the IEEE/CVF Conference on Computer Vision and Pattern Recognition Workshops (CVPRW)},
  year={2024}
}

@inproceedings{lim2025ihalla,
  title={I-HallA: Image Hallucination Evaluation with Question Answering},
  author={Lim, Youngsun and Choi, Hojun and Shim, Hyunjung},
  booktitle={Proceedings of the AAAI Conference on Artificial Intelligence (AAAI)},
  year={2025}
}
}
\newpage
\appendix

\end{document}